\documentclass{article}

\PassOptionsToPackage{numbers, compress}{natbib}

\usepackage[preprint]{neurips_2024}

\usepackage[utf8]{inputenc}
\usepackage[T1]{fontenc}
\usepackage{url}
\usepackage{booktabs}
\usepackage{amsfonts}
\usepackage{nicefrac}
\usepackage{microtype}
\usepackage{xcolor}

\usepackage{wrapfig}
\usepackage{subcaption}

\usepackage{color, colortbl}

\usepackage{algorithm}
\usepackage{listings}
\usepackage{caption}
\usepackage{float}
\definecolor{DGray}{gray}{0.45}
\definecolor{Gray}{gray}{0.9}
\definecolor{codeblue}{rgb}{0.25,0.5,0.5}
\definecolor{codekw}{rgb}{0.85, 0.18, 0.50}
\definecolor{codekwb}{rgb}{0.0, 0.0, 1.00}
\floatstyle{plain}
\newfloat{Code}{htbp}{Code}
\restylefloat*{Code}
\floatname{Code}{Code}

\usepackage[pagebackref=true,breaklinks=true,letterpaper=true,colorlinks,bookmarks=false]{hyperref}
\usepackage{amsmath}
\usepackage[capitalize]{cleveref}
\crefname{section}{Sec.}{Secs.}
\Crefname{section}{Section}{Sections}
\Crefname{table}{Table}{Tables}
\crefname{table}{Tab.}{Tabs.}

\usepackage{pifont}

\usepackage{times}
\usepackage{epsfig}
\usepackage{graphicx}
\usepackage{amssymb}

\usepackage{multirow}
\usepackage{booktabs}
\usepackage{float}

\usepackage{bbding}

\newcommand{\eg}{\textit{e.g.}}
\newcommand{\ie}{\textit{i.e.}}

\title{DiM: Diffusion Mamba for \\Efficient High-Resolution Image Synthesis}

\author{
  Yao Teng\textsuperscript{1} \quad Yue Wu\textsuperscript{2} \quad Han Shi\textsuperscript{2}  \quad Xuefei Ning\textsuperscript{3} \\ \bf{Guohao Dai}\textsuperscript{4}  \quad \bf{Yu Wang}\textsuperscript{3} \quad \bf{Zhenguo Li}\textsuperscript{2}  \quad \bf{Xihui Liu}\textsuperscript{1}\thanks{Corresponding Author} \\
\textsuperscript{1}The University of Hong Kong  \quad
\textsuperscript{2}Huawei Noah’s Ark Lab \\ \quad \textsuperscript{3}Tsinghua University \quad
\textsuperscript{4}Shanghai Jiao Tong University\\
}

\begin{document}

\maketitle

\begin{abstract}
  Diffusion models have achieved great success in image generation, with the backbone evolving from U-Net to Vision Transformers. However, the computational cost of Transformers is quadratic to the number of tokens, leading to significant challenges when dealing with high-resolution images. In this work, we propose Diffusion Mamba (DiM), which combines the efficiency of Mamba, a sequence model based on State Space Models (SSM), with the expressive power of diffusion models for efficient high-resolution image synthesis. To address the challenge that Mamba cannot generalize to 2D signals, we make several architecture designs including multi-directional scans, learnable padding tokens at the end of each row and column, and lightweight local feature enhancement. Our DiM architecture achieves inference-time efficiency for high-resolution images. In addition, to further improve training efficiency for high-resolution image generation with DiM, we investigate ``weak-to-strong'' training strategy that pretrains DiM on low-resolution images ($256\times 256$) and then finetune it on high-resolution images ($512 \times 512$). We further explore training-free upsampling strategies to enable the model to generate higher-resolution images (\eg, $1024\times 1024$ and $1536\times 1536$) without further fine-tuning. Experiments demonstrate the effectiveness and efficiency of our DiM.
  The code of our work is available here: {\url{https://github.com/tyshiwo1/DiM-DiffusionMamba/}}.
\end{abstract}

\section{Introduction}
\label{sec:intro}

\begin{figure}
    \centering
    \includegraphics[width=0.9\linewidth]{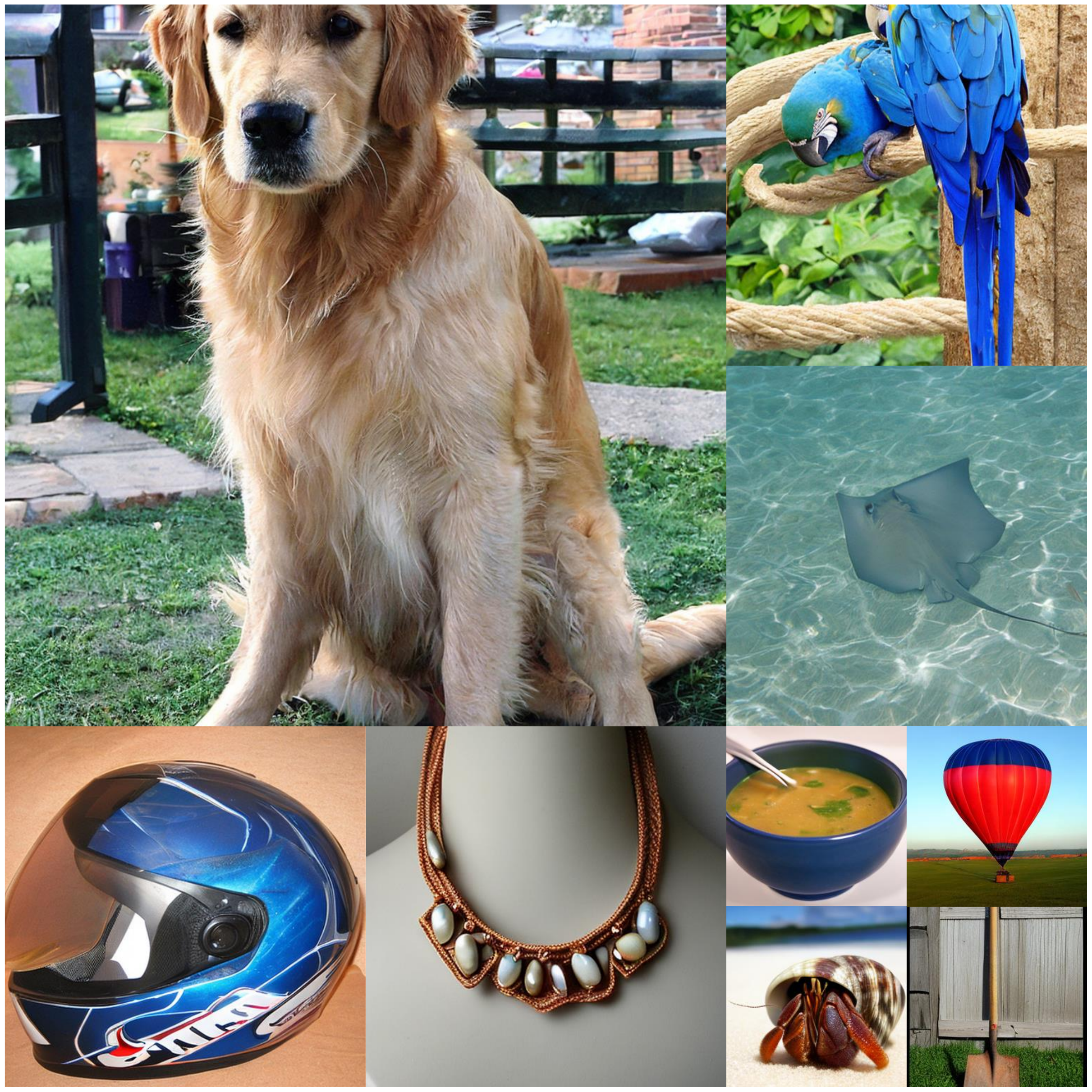}
    \caption{
    The images generated by our model trained on ImageNet.
    The resolutions are $1024 \times 1024$, $512 \times 512$ and $256 \times 256$.
    The classifier-free guidance weight is set to 4.0.
    }
    \label{fig:teaser}
    \vspace{-2em}
\end{figure}

Diffusion models have achieved great success in image generation~\cite{ddpm,ddim,Dalle-2,stable_diffusion}.
The backbones of diffusion models have evolved from Convolutional Neural Networks represented by U-Net~\cite{unet} to Vision Transformers~\cite{dit,uvit,pixartalpha,sd3,sora}, because of the effectiveness and scalability of transformer architectures. Transformer-based diffusion models encode images into latent feature maps, patchify the latent feature maps, project patches to tokens, and then apply transformers to denoise the image tokens.
However, the complexity of self-attention layers in transformers is quadratic to the number of tokens, leading to substantial challenges in terms of computation cost for high-resolution generation.

Recently, Mamba~\cite{mamba}, a sequence model backbone based on State Space Models~(SSM)~\cite{ssm_graduate}, has shown remarkable effectiveness and efficiency across several modalities such as language, audio, and genomics~\cite{mamba,mambaspeech}. Mamba achieves comparable performance with Transformers with better inference time efficiency. In particular, Mamba shows great promise in long sequence modeling, because the computational complexity of Mamba scales linearly with the number of tokens, compared to quadratic scaling for Transformers. Those properties of Mamba motivate us to introduce Mamba as a new backbone into diffusion models, particularly for efficient high-resolution image generation.

However, several challenges arise when incorporating Mamba with diffusion models for high-resolution image generation. The primary challenge stems from the mismatch between the causal sequential modeling of Mamba and the two-dimensional (2D) data structure of images. Mamba is designed for one-dimensional (1D) causal modeling for sequential signals, which cannot be directly leveraged for modeling two-dimensional image tokens. A simple solution is to use the raster-scan order to convert 2D data into 1D sequences. However, it restricts the receptive field of each location to only the previous locations in the raster-scan order. Moreover, in the raster-scan order, the ending of the current row is followed by the beginning of the next row, while they do not share spatial continuity. The second challenge is, despite the advantage of Mamba for efficient inference, training Mamba-based diffusion models on high-resolution images is still costly.

\begin{figure*}
    \centering
    \includegraphics[width=0.9\linewidth]{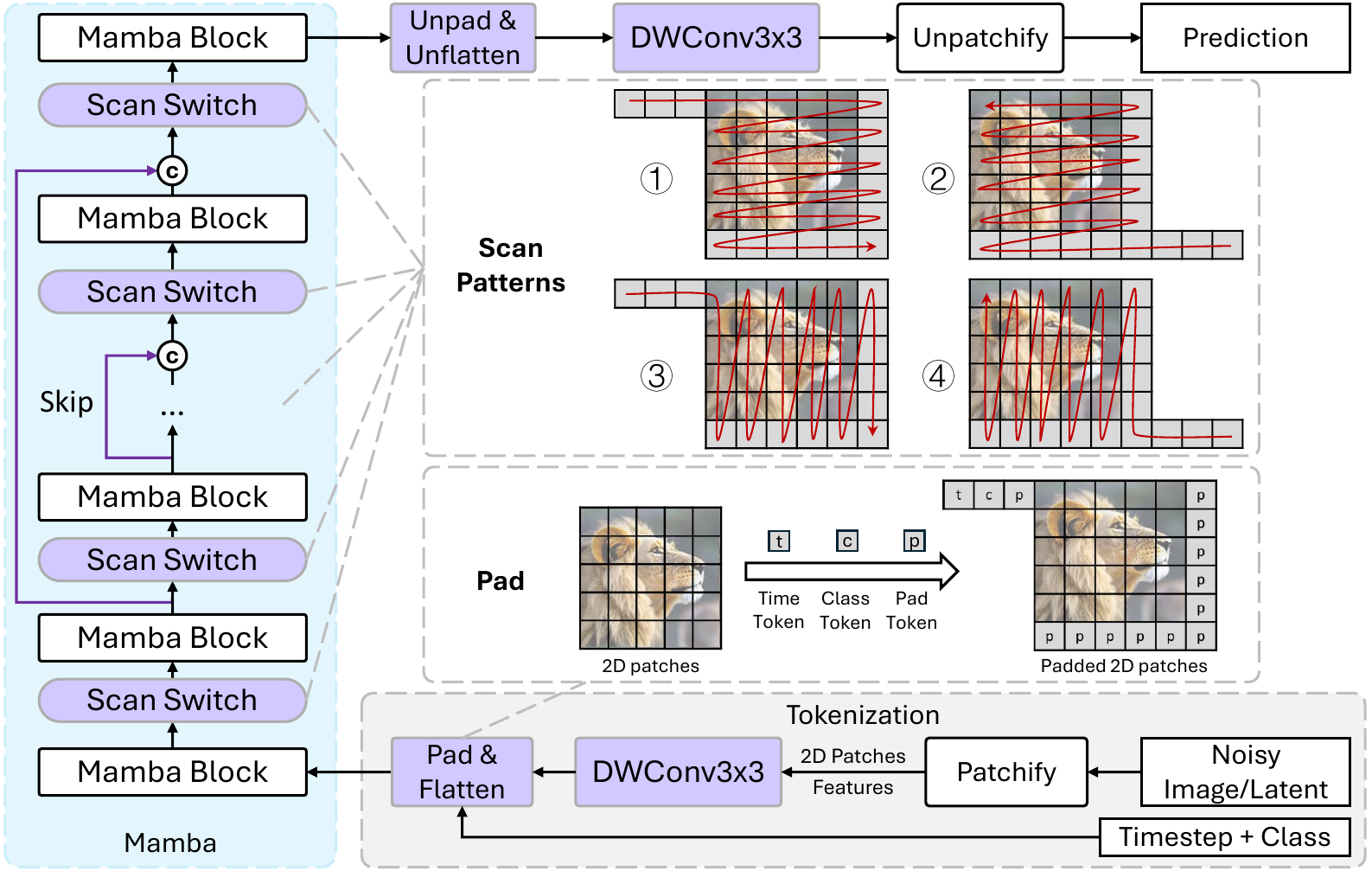}
    \caption{
    \textbf{Overview.} 
    The inputs of our framework is a noisy image/latent, with a timestep and a class condition.
    The noisy inputs are transformed into patch-wise features, processed by a depth-wise convolution, and appended with time, class, and padding tokens.
    The features are flattened and scanned by Mamba blocks with four directions.
    The features are then transformed into 2D patches, processed by another convolution, and finally used for noise prediction. 
    }
    \label{fig:main}
\end{figure*}

To mitigate the first challenge, we propose Diffusion Mamba (DiM) (shown in~\cref{fig:main}), a Mamba-based diffusion model backbone for efficient high-resolution image generation. 
In our framework, we follow transformer-based diffusion models to encode and patchify images into patch features. 
Then, we leverage Mamba architecture as the diffusion backbone to model the patch features.
In order to avoid unidirectional causal relationships among patches and to endow each token with a global receptive field, 
we design the Mamba blocks to alternately execute four scanning directions.
Moreover, we insert learnable padding tokens between the two tokens which are adjacent in the scanning order but not adjacent in the spatial domain,
so as to allow the Mamba blocks to discern the image boundaries and to avoid misleading the sequence model. 
We also add $3 \times 3$ depth-wise convolution layers to the input layer and output layer of the network to enhance the local coherence of generated images.
Additionally, we add long skip connections~\cite{unet,uvit} between shallow and deep layers to propagate the low-level information to the high-level features, which is proven to be beneficial for the pixel-level prediction objective in diffusion models. 

To tackle the challenge of training efficiency for high-resolution image generation with DiM, we explore resource-efficient approaches to adapt our DiM model pretrained on low-resolution images for high-resolution image generation. 
We first observe empirical evidence that our DiM pretrained on low-resolution images provides reasonable prior for high-resolution image generation.
So we explore the ``weak-to-strong'' training strategy~\cite{pixartsigma} where we first train DiM on low-resolution images, and then use the pretrained model as initialization to efficiently fine-tune on high-resolution images. This strategy largely reduces the training time cost for high-resolution image generation.
We also explore training-free upsampling approaches to adapt DiM to generate higher-resolution images without further finetuning.

We conduct experiments on CIFAR-10~\cite{cifar} and ImageNet~\cite{imagenet}.
On CIFAR-10, DiM-Small can achieve an FID score of 2.92, and we also perform ablation studies to demonstrate the effectiveness of our designs.
On ImageNet, we pretrain our DiM-Huge at a resolution of $256 \times 256$, and then fine-tune DiM-Huge efficiently at a higher resolution of $512 \times 512$. 
Despite trained with much fewer iterations, our DiM-Huge achieves comparable performance to other transformer-based and SSM-based diffusion models, demonstrating the effectiveness and training efficiency of our approach. Moreover, with training-free upsampling schemes, our DiM fine-tuned on $512 \times 512$ images can further generate $1024 \times 1024$ and $1536 \times 1536$ images. We analyze the inference time to demonstrate the efficiency of DiM for high-resolution image synthesis.

In summary, our contributions lie in the following aspects: 
\begin{itemize}
 \item We propose a new Mamba-based diffusion model, DiM, for efficient high-resolution image generation. We propose several effective designs to endow Mamba, which was designed for processing 1-D signals, with the ability to model 2-D images. 
 \item To address the high cost of training on high-resolution images, we investigate strategies to fine-tune DiM pretrained on low-resolution images for high-resolution image generation. Moreover, we explore training-free upsampling schemes to adapt the model to generate higher-resolution images without further fine-tuning.
 \item The experiments on ImageNet and CIFAR demonstrate the training efficiency, inference efficiency, and effectiveness of our DiM in high-resolution image generation.
\end{itemize}
\section{Related Work}
\label{sec:rela}

\noindent
\textbf{Backbones of Diffusion Models.}
Traditionally, U-Net~\cite{unet} serves as a backbone for diffusion models~\cite{ddpm,stable_diffusion,sdxl}.
This architecture is characterized by its down-sampling and up-sampling blocks, connected by long skip connections.
Each block of this U-Net is composed of convolutional layers and attention modules~\cite{transformer}.
However, the scalability of this architecture has not been successfully demonstrated.
Recently, transformer-based diffusion models have been proposed~\cite{dit,pixartalpha,sd3,uvit,lumina-t2x}.
Different from the diffusion models based on U-Net, the architecture of these transformer-based models is built exclusively on attention modules and multi-layer perceptrons (MLPs), without any convolutional layers.
These models have shown remarkable scalability in the generative tasks of computer vision~\cite{lumina-t2x,hunyuandit,sora}.
Notably, Sora~\cite{sora} exemplifies the great scalability of transformers in generating high-quality videos. 
To effectively train these diffusion models with scaled-up backbones, many works suggest performing the training process at lower resolutions and then fine-tuning the pre-trained models at higher resolutions~\cite{sd3,pixartalpha,palix}.
While the transformer excels in general image generation, its limited efficiency in processing large quantities of tokens hinders the progression of high-resolution image generation.

\noindent
\textbf{State Space Models.}
State Space Model~(SSM)~\cite{ssm_graduate,ssm_h,ssm_s4,ssm_h3} is proposed for sequential modeling, and it has been applied in control theory, signal processing, and natural language processing~\cite{ssm_graduate}.
The inputs and outputs of SSMs are one-dimensional sequences.
In the process of SSMs, the tokens from the input sequence are recurrently mapped to the hidden states through a linear transformation~(addition and element-wise multiplication) with the preceding hidden states and the model weights.
The outputs are also originated from the hidden states via linear transformation with other model weights.
Mamba~\cite{mamba} is a new type of state space model where each block contains a selective scan module, a 1D causal convolution, and a normalization layer.
The process of the selective scan is similar to that in SSMs, but it implements the function of data selection via input-dependent model weights~\cite{dynamic_filter}.
To accelerate the training and inference of Mamba, this selective scan is implemented via a work-efficient parallel scanning algorithm~\cite{selective_scan1} in SRAM.
In the following paragraph, we will delve into the existing state space models used for image generation and provide a detailed comparison with our method.

\noindent
\textbf{State Space Models in Vision.}
The state space models have already been applied in computer vision even prior to Mamba. DiffuSSM~\cite{diffussm} is the first diffusion model replacing attention mechanisms with state space models.
Recently, Mamba has been proposed for better modeling power compared to the preceding state space models.
Various Mamba variants are then proposed for vision tasks on image and video inputs~\cite{vim,vmamba,videomamba1,videomamba2}.
Also, several concurrent works introduce Mamba into image generation.
DiS~\cite{dis} directly incorporates ViM~\cite{vim}, a variant of Mamba, into image generation, exploring its generative capabilities up to a maximum of $512 \times 512$ resolution images.
ZigMa~\cite{zigma} utilizes the vanilla Mamba blocks with various scan patterns and is trained on high-resolution human face generation datasets~\cite{ffhq}.
% However, the performance of this method on the natural objects is not comparable to other transformer-based diffusion models such as U-ViT~\cite{uvit}.
% not verify 
% More importantly, all of these works choose to train their models from scratch on each dataset.
Distinct from the above methods,  
we propose the first Mamba-based diffusion model which can generate images with more than 10K tokens, unleashing the power of Mamba on long sequence processing.
Our model also contains several new modules for spatial prior and achieves comparable performance to the transformer-based diffusion models on the widely used benchmarks.
We are also the first framework to validate the \textit{fine-tuning} ability of Mamba-based diffusion model at various resolutions.

\section{Method}
\label{sec:method}

We introduce the preliminaries of State Space Models and Mamba in Sec.~\ref{sec:3.1}. We then introduce the network architecture and several designs of our Mamba-based diffusion model (DiM) in Sec.~\ref{sec:3.2}. The several designs of DiM adapts Mamba, which was designed for processing 1-D signals, for 2D image generation, and enables \textit{efficient inference} for high-resolution image generation. In Sec.~\ref{sec:3.3}, we investigate the fine-tuning and training-free strategies to improve the \textit{training efficiency} of DiM for high-resolution image generation.

\subsection{Preliminaries of State Space Models}\label{sec:3.1}

State Space Models (SSM)~\cite{ssm_graduate,ssm_h,ssm_h3,ssm_s4} are designed to encode and decode one-dimensional sequential inputs.
In a continuous-time SSM, an input signal $x(i)$ is first encoded into a hidden state vector $h(i)$ and then decoded into the output signal ${y}(i)$ according to the following ordinary differential equations (ODEs):
\begin{align}
    {h}'(i) = \mathbf{A}~{h}(i) + \mathbf{B}~x(i), ~~~~~ {y}(i) =  \mathbf{C}~{h}(i) + \mathbf{D}~x(i),
\end{align}
where ${h}'$ denotes the derivative of $h$, and $\mathbf{A}$, $\mathbf{B}$, $\mathbf{C}$, $\mathbf{D}$ denote the weights of SSMs.
Typically, the inputs in natural languages and two-dimensional vision are discrete signals, so Mamba leverages the zero-order hold (ZOH) rule for discretization.
Thus, the above ODEs can be recurrently solved:
\begin{align}
    &\mathbf{\bar{A}} = \exp \left( \mathbf{\Delta} \mathbf{A} \right), ~~ \mathbf{\bar{B}} = \left( \mathbf{\Delta} \mathbf{A} \right)^{-1} \left( \exp \left( \mathbf{\Delta} \mathbf{A} \right) -\mathbf{I} \right) \cdot \mathbf{\Delta} \mathbf{B} , \\
    &{h}_{i} = \mathbf{\bar{A}}~{h}_{i-1} + \mathbf{\bar{B}}~{x}_{i}, ~~~~ {y}_{i} = \mathbf{C}~{h}_{i} + \mathbf{D}~{x}_{i},
\end{align}
where $\mathbf{\Delta}$ is also a model parameter. 
Recently, Mamba~\cite{mamba} proposes to improve the flexibility of SSM by changing the time-invariant parameters to be time-varying.
This modification involves replacing the static model weights $(\mathbf{B}, \mathbf{C}, \mathbf{\Delta})$ with the dynamic weights~\cite{dynamic_filter} dependant on the input $x$.
This process with input-dependant parameters is termed as the selective scan.

\subsection{Architecture of Diffusion Mamba (DiM)}\label{sec:3.2}

Mamba is primarily designed for processing one-dimensional inputs, so it is difficult for Mamba to learn the two-dimensional data structure of images without any modification.
Therefore, we propose several new architectural designs that enable DiM to handle spatial structure.

\textbf{Overall architecture.} As depicted in~\cref{fig:main}, our framework processes a noisy two-dimensional (2D) input, such as an image or latent features~\cite{stable_diffusion}, with a timestep and a class condition.
This noisy input can be deemed as a clean signal perturbed by a certain level of Gaussian noise corresponding to the input timestep.
The noisy input is first split into 2D patches, and each patch is transformed into a high-dimentional feature vector by a fully-connected layer.
Next, these patches are fed into a $3 \times 3$ depth-wise convolution layer, where the local information is injected into the patches.
The patches are also padded with learnable tokens at the end of rows and columns, allowing the model to be aware of the 2D spatial structure during the 1-D sequential scanning.
Then, the patch tokens are flattened into a \textit{patch sequence}, using one of the four scan patterns illustrated in~\cref{fig:main}.
The timestep and class condition are also transformed into tokens by fully-connected layers, and are then appended to the sequence~\cite{uvit}.
Subsequently, the sequence is fed into the Mamba blocks for scanning. 
Additionally, we add long skip connections~\cite{unet,uvit} between shallow and deep layers to propagate the low-level information to the high-level features, which is proven to be beneficial for the pixel-level prediction objective in diffusion models. We illustrate several design choices in the following paragraphs.

\noindent
\textbf{Scan patterns.}
A global receptive field is crucial for our model to efficiently capture the spatial structure within images.
Scanning the image patches in a single raster-scan direction leads to uni-directional and limited receptive fields of patches.
For example, the first scanned patch on the top-left corner would never aggregate the information from other patches.
To allow each patch to have a global receptive field we adopt different scanning patterns at different model blocks.
Specifically, as shown in~\cref{fig:main}, in the first block, we adopt the row-major scan, \ie, we scan the sequence of image patches row by row, with each row being scanned horizontally from left to right and then move to the next row. In the second block, we reverse the sequence order and scan the sequence in the same manner. 
In the subsequent blocks, we perform column-major scans in both the forward and reverse order.
After traversing all scan patterns, we loop over them again across the next model blocks.

\noindent
\textbf{Learnable padding token.}
The learning of spatial structure of images may be disrupted by the raster scan.
To be specific, when we flatten an image into a patch sequence, the right-most patch in one row of the image becomes adjacent to the left-most patch of the next row. 
However, the contents represented by these two feature vectors may vastly differ. This contradicts the inherent continuity and spatial structure of images, thereby hindering the learning process.
To mitigate this issue, we allow the model to be aware of the end-of-line~(EOL) by appending learnable padding tokens at the end of each row or each column.

\noindent
\textbf{Lightweight local feature enhancement.}
The local structure of images is disrupted by the flattening of tokens for the scan.
For example, in the row-major scan, the patch at row $i$ and column $j$ is no longer adjacent to the patch at row $(i+1)$ and column $j$.
Furthermore, since Mamba has been designed for extreme efficiency, we opt to enhance the local structure by adding a few lightweight modules at the beginning and the end of the network, instead of altering the Mamba blocks.
To be specific, we introduce two $3 \times 3$ depth-wise convolution layers.
One convolution layer is inserted after the patchify layer before feeding the tokens into Mamba blocks. Another convolution layer is inserted after all the Mamba blocks, before the unpatchify and output layer. Those lightweight depth-wise convolution layers provide DiM with awareness of the 2D local continuity.

\subsection{Training and Inference Strategies}\label{sec:3.3}

Despite its \textbf{inference-time efficiency}, training DiM on high-resolution images requires a lot of time and computational resources. In this subsection, we investigate the strategies to improve the \textit{training efficiency} of DiM for high-resolution images.

\noindent
\textbf{``Weak-to-strong'' training and fine-tuning.}
Training a diffusion model from scratch for high-resolution images requires a lot of time and computational resources.
We observe that DiM pretrained on low-resolution images can provide a rough initialization for high-resolution training, shown in~\cref{fig:direct512_init}.
Therefore, we consider a ``weak-to-strong'' training strategy~\cite{pixartsigma} where we pretrain our model from scratch on low-resolution images and then perform fine-tuning on higher resolutions.
During fine-tuning, we upscale the length and width of images by a factor of 2.
This strategy largely reduces the computational cost for training high-resolution image generators with DiM.

\noindent
\textbf{Training-free upsampling.}
Extremely high-resolution images with annotations are not easy to obtain, making it difficult to fine-tune to DiM to higher resolution.
Therefore, we explore the training-free resolution upsampling capability of our model.
For example, we directly use our model trained on $512 \times 512$ dataset~\cite{imagenet} to generate $1024 \times 1024$ images.
However, performing training-free super-resolution image generation on our model is non-trivial.
We observe that directly feeding our network with a higher-resolution Gaussian noise results in images with repetitive patterns, corrupted global structures, and collapsed spatial layouts. 
Only the local structure and details exhibit relatively good quality.
In order to generate better global structures, we utilize the upsample-guidance~\cite{upsample_guidance} at the early diffusion timesteps (\eg, the first $30\%$ of timesteps):
\begin{equation}
    \epsilon_{\theta} (x_t, t) = \epsilon_{\theta} (x_t, t) + \omega_t \mathbf{U} \left[ \frac{1}{m} \epsilon_{\theta} \left( \frac{ 1 }{ \sqrt{ P_t }  } \mathbf{D}\left[ x_t \right] , \tau \right) - \mathbf{D}\left[ \epsilon_{\theta} (x_t, t) \right] \right], \\
\end{equation}
where $\epsilon_{\theta}$ denotes our noise-prediction Mamba-based model, 
$m$ denotes the upscaling factor~(\eg, $m=2$),
$\mathbf{U}$ denotes the nearest upsampling operator by scale $m$, 
$\mathbf{D}$ denotes the average pooling operator (down-sampling) by stride $m$, 
$x_t$ denotes a noisy input,
$t$ denotes an input diffusion timestep,
$\tau$ denotes the timestep whose signal-to-noise ratio is $m^2$ times the signal-to-noise ratio of $t$,
$P_t$ denotes an coefficient to calibrate the overall power of the predicted noise at each timestep, and
$\omega_t$ denotes the weight for upsample-guidance.
In the later diffusion timesteps, we directly feed the higher-resolution noisy inputs into DiM for noise prediction.
\section{Experiments}
\label{sec:experiments}

\subsection{Experimental Setup}

\begin{wraptable}{r}{19em} 
\vspace{-1em}
  \small
  \setlength{\tabcolsep}{3pt}
  \caption{The configuration of our model.}
  \vspace{-.5em}
  \label{tab:model_size}
  \centering
  \begin{tabular}{l|cccc}
    \toprule
 Model & Params & Blocks & Hidden dim & Gflops \\
 \midrule
 Small (S) & 50M &  25 & 512 & 12 \\
 Large (L) & 380M & 49 & 1024 & 94 \\
 Huge (H) & 860M & 49 & 1536 & 210 \\
  \bottomrule
  \end{tabular}
\end{wraptable}

\textbf{Model configuration.}
Following the existing settings~\cite{uvit,dis,zigma}, we present three versions of our framework with different model sizes in~\cref{tab:model_size}, where the Gflops is calculated with a batch size of 1, and the input size is set as $32 \times 32$ without image autoencoder~\cite{ae}.
As for the hyper-parameters of Mamba blocks, we follow the standard settings~\cite{mamba}.
Following the traditional settings~\cite{uvit,dit}, we set the patch size as $2 \times 2$ for DiM trained on ImageNet and CIFAR.

\textbf{Implementation details.}
Every \textit{training} experiment is performed on 8~A100-80G.
Following the previous works~\cite{uvit}, we use the same DDPM~\cite{ddpm} scheduler, pretrained image autoencoder~\cite{ae} and DPM-Solver~\cite{dpm_solver}. We use random flip as the data augmentation.
The learning rate is set to $2 \times 10^{-4}$. We also use EMA with a rate of 0.9999.

\textbf{Datasets and evaluation metrics.}
We use FID-50K~\cite{fid} as the metrics on all datasets for evaluation.
The specific settings for each dataset are as follows:
(1)~CIFAR: The model is trained for unconditional image generation with a batch size of 128.
(2)~ImageNet: We train the model for conditional image generation. We also use classifier-free guidance for evaluation, and the guidance weight for calculating FID is identical to that in~\cite{uvit}. When performing pretraining on ImageNet $256 \times 256$, we set the batch size as 1024 and 768 for DiM-Large and DiM-Huge, respectively.
When fine-tuning DiM-Huge on ImageNet $512 \times 512$, we set the batch size as 240 with gradient accumulation.

\textbf{Training and inference setups.} 
On ImageNet, we first pretrain DiM with more than 300K iterations at $256 \times 256$ resolution.
We then finetune the pretrained model on $512 \times 512$ resolution.
To achieve a higher resolution without the cost of training, we further use training-free upsampling techniques to generate $1024 \times 1024 $ and $1536 \times 1536$ images with DiM-Huge trained on $512 \times 512$ resolution.

\subsection{Efficiency Analysis}

In this subsection, we examine the efficiency of DiM and compare it the transformer backbone.
A single selective scan is more efficient than FlashAttention~V2~\cite{flashattn2}.
However, to maintain a similar number of parameters, the standard Mamba has twice as many blocks as the transformer.
\begin{wrapfigure}{r}{18em} 
\centering
\includegraphics[width=0.99\linewidth]{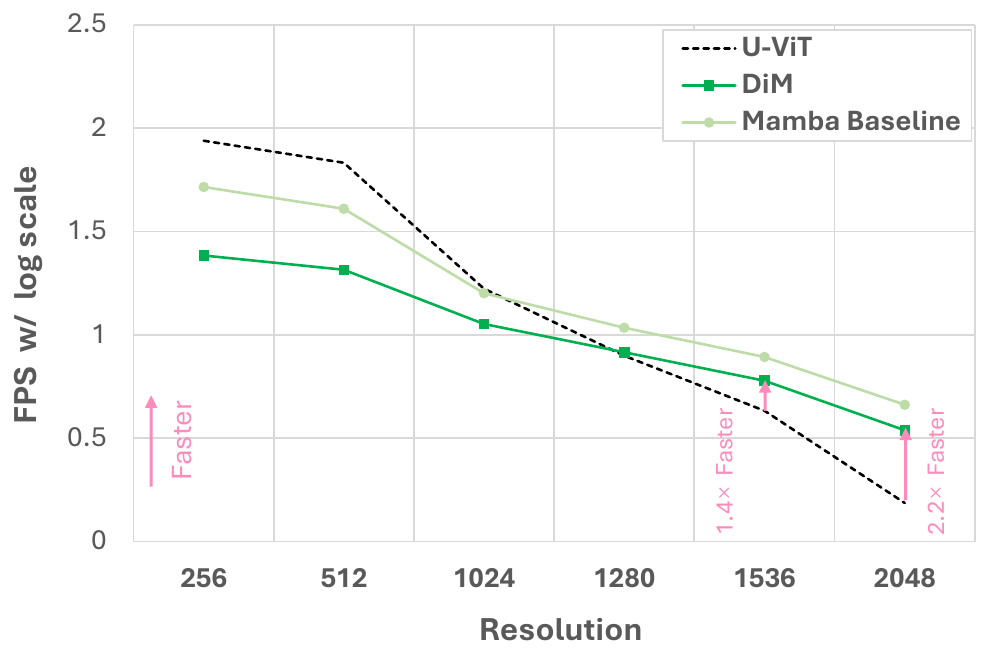}
\caption{
The inference speed across models with varied resolutions. Each model has 0.9 billion parameters. We present the speed as the logarithm of iterations per millisecond for clearer visualization.
}
\label{fig:infer_speed}
\end{wrapfigure}
These doubled scans increase the computational complexity.
Additionally, our proposed modules including the switching of scanning patterns also create slight latency.
To compare the practical efficiency of the transformer and Mamba on image generation, 
we perform experiments on a single H800 GPU.
We present the inference speed of our model, U-ViT~\cite{uvit} and a Mamba baseline in~\cref{fig:infer_speed}. These models have a similar parameter count (0.9B) and the same $2 \times 2$ patch size. We can see that the original Mamba baseline and DiM are slower than the well-optimized transformer-based models at resolution below $1024 \times 1024$. Nevertheless, DiM is faster than the transformer at resolutions above $1280 \times 1280$, thanks to its linear complexity.
Furthermore, our DiM is only slightly less efficient than the Mamba baseline, indicating that our designs added to the original Mamba adapts Mamba to process 2D images but do not cause large additional computational cost.
We conduct experiments to generate images at a resolution of $ 1536 \times 1536$ and present the results in~\cref{fig:imagenet1536}. This demonstrates that DiM can generate meaningful high-resolution images faster than transformer.

\subsection{Qualitative Results}

\begin{figure}
    \begin{subfigure}[h]{0.66\textwidth}
        \includegraphics[width=1\linewidth]{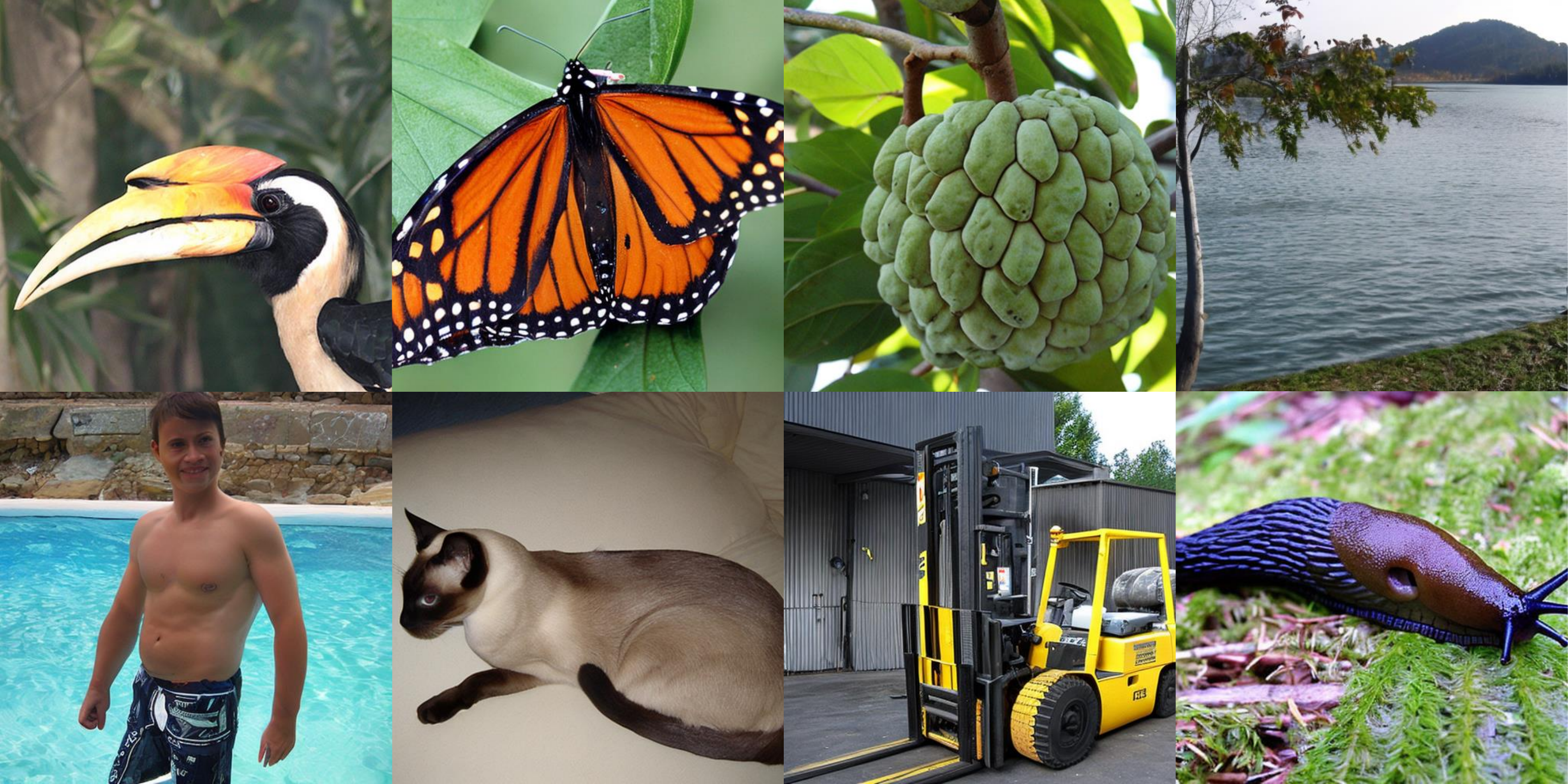}
        \caption{ImageNet $512 \times 512$}
        \label{fig:imagenet512cfg}
    \end{subfigure}
    \begin{subfigure}[h]{0.33\textwidth}
        \includegraphics[width=1\linewidth]{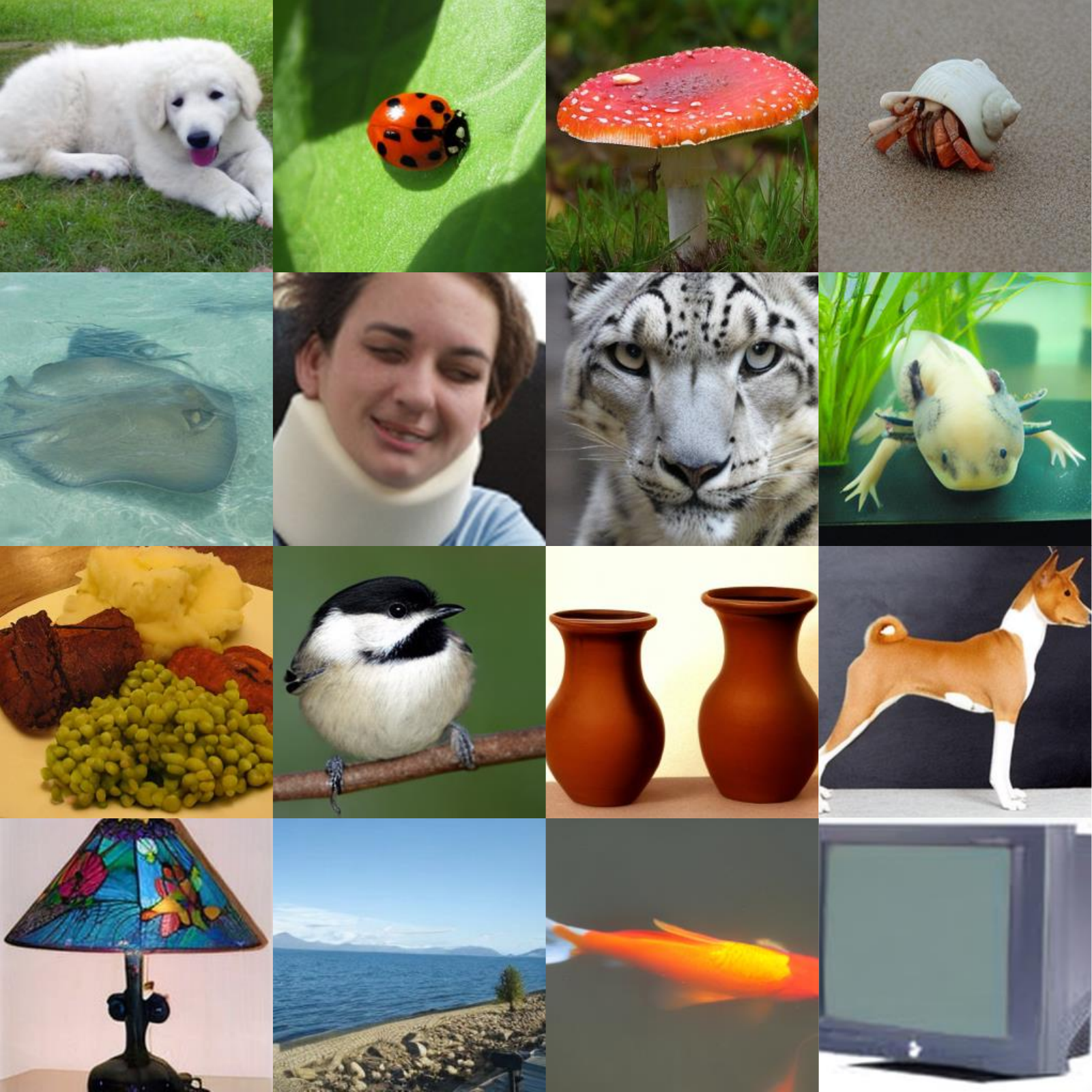}
        \caption{ImageNet $256 \times 256$}
        \label{fig:imagenet256cfg}
    \end{subfigure}
\caption{The images generated by DiM-Huge with \texttt{cfg=4.0}.}
\vspace{-1em}
\end{figure}

\begin{figure}
    \begin{subfigure}[h]{0.66\textwidth}
        \includegraphics[width=1\linewidth]{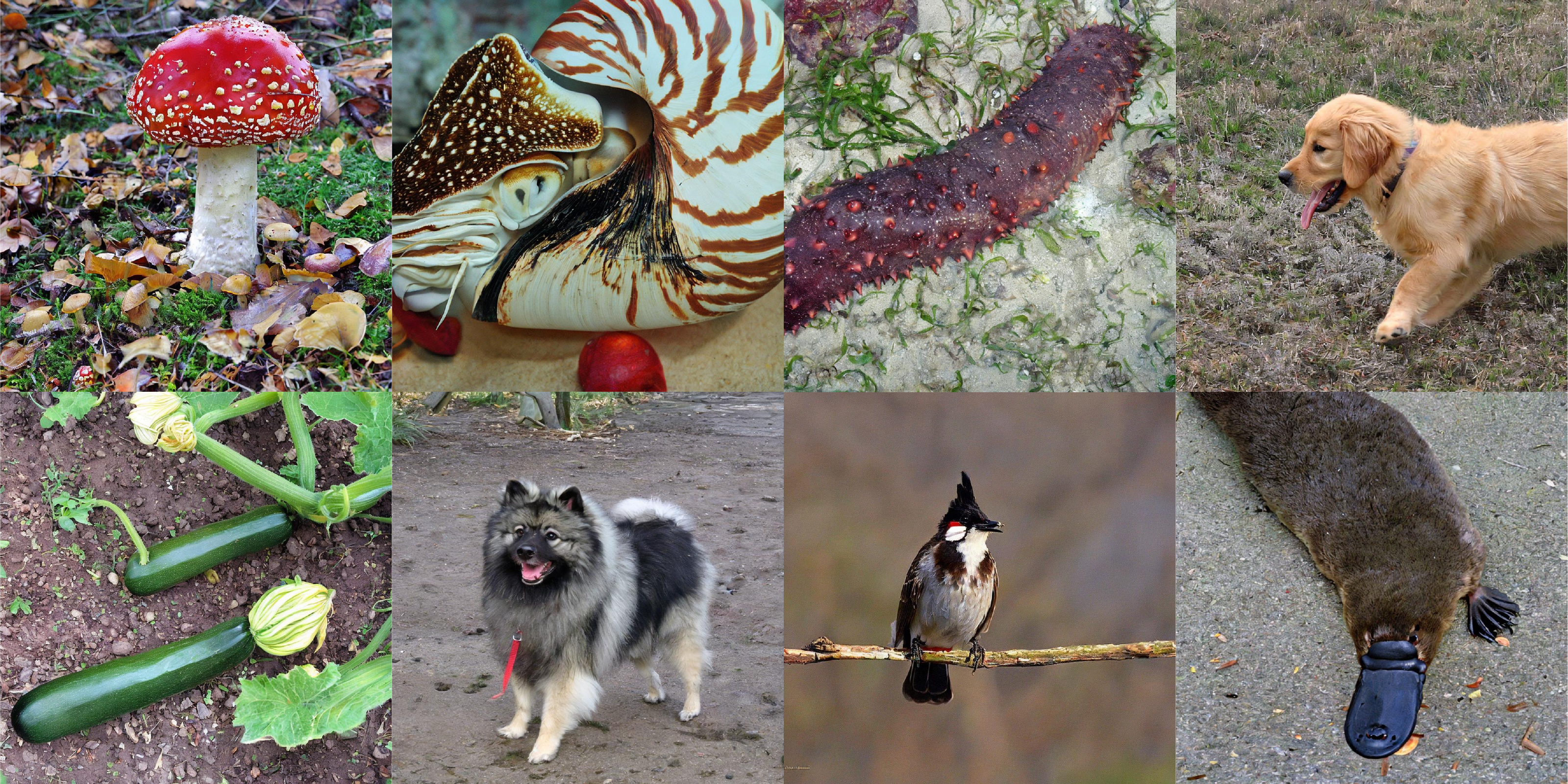}
        \caption{ImageNet $1024 \times 1024$}
        \label{fig:imagenet1024}
    \end{subfigure}
    \begin{subfigure}[h]{0.33\textwidth}
        \includegraphics[width=1\linewidth]{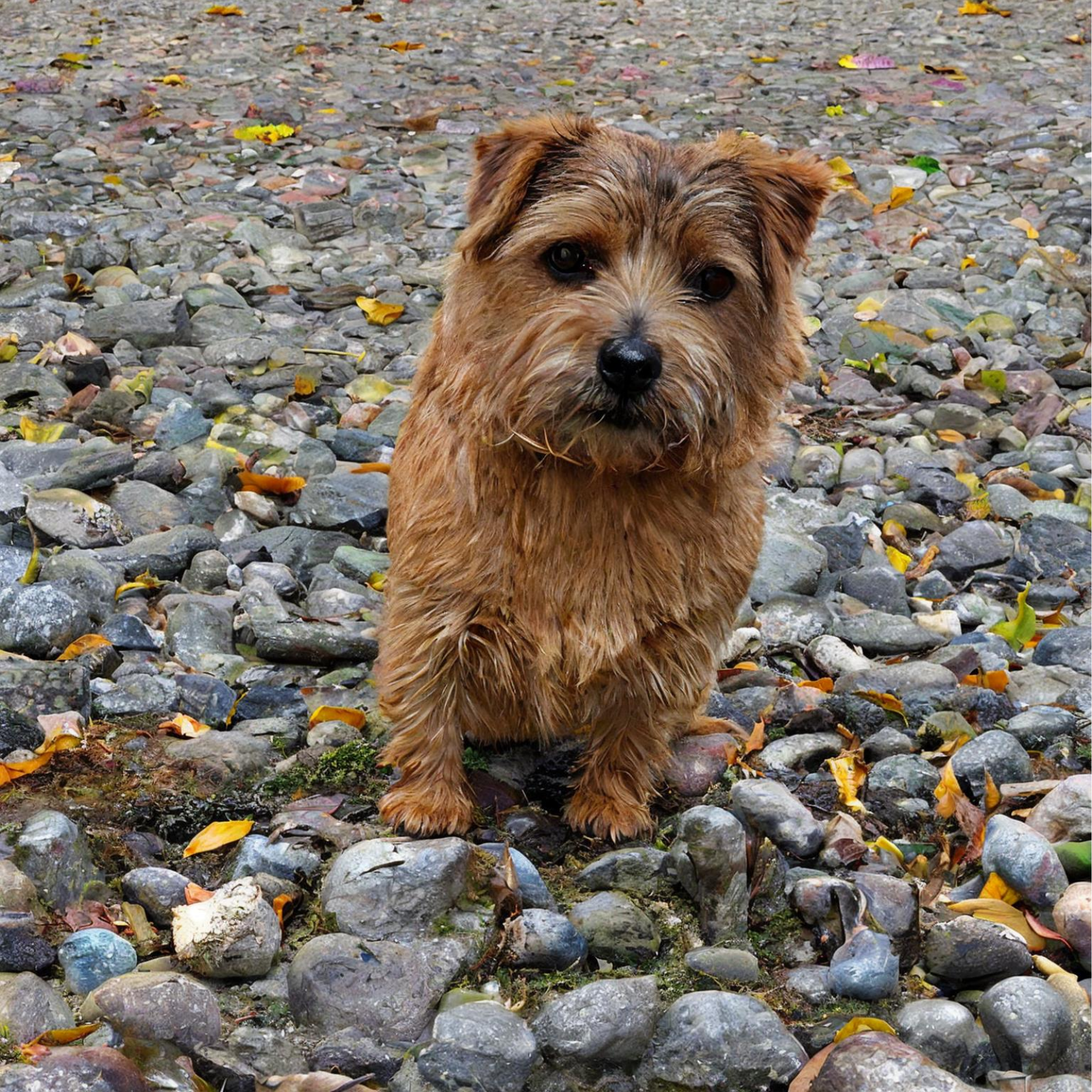}
        \caption{ImageNet $1536 \times 1536$}
        \label{fig:imagenet1536}
    \end{subfigure}
\caption{The high-resolution images generated by DiM-Huge trained on $512 \times 512$ images.}
\label{fig:ug_imagenet}
\vspace{-1em}
\end{figure}

\noindent
\textbf{Model trained on ImageNet.} As shown in~\cref{fig:imagenet256cfg}, we select a set of generated images for visualization. The results show that DiM-Huge pretrained on ImageNet can generate high-quality $256 \times 256$ images with a classifier-free guidance weight of 4.0.
As shown in~\cref{fig:imagenet512cfg}, our model fine-tuned on ImageNet with $512 \times 512$ resolution also shows great performance.

\textbf{Training-free up-sampling.}
We can use our model trained on $512 \times 512$ ImageNet dataset~\cite{imagenet} to directly generate $1024 \times 1024$ and $1536 \times 1536$ images.
As shown in~\cref{fig:ug_imagenet}, even when the resolution is increased to three times that of training, our model is still able to generate visually appealing images with upsample-guidance~\cite{upsample_guidance}. 

\begin{table}[tb]
  \caption{Pretraining on ImageNet with the resolution $256 \times 256$. }
  \label{tab:imagenet256}
  \centering
  \small
  \begin{tabular}{l|c|c|c|c}
  \toprule
 Mode & Images~(Iterations $\times$ BatchSize) & Parameters & GFlops & FID \\
 \midrule
 U-ViT-L/2~\cite{uvit} & 300M~(300K $\times$ 1024) & 287M & 77 & 3.52 \\
 U-ViT-H/2~\cite{uvit} & 500M~(500K $\times$ 1024) & 501M & 133 & 2.29 \\
 DiT~\cite{dit}  & 1792M~(7M $\times$  256) &  675M & 119 & 2.27 \\
 DiffuSSM-XL~\cite{diffussm} & 660M & 660M & 280 & 2.28 \\ 
 \midrule
 \textbf{DiM-Large} & 300M~(300K $\times$ 1024) & 380M & 94 & 2.64 \\ 
 \textbf{DiM-Huge} & 319M~(425K $\times$ 768) & 860M & 210  &  2.40 \\
 \textbf{DiM-Huge} & 480M~(625K $\times$ 768) & 860M & 210  &  \textbf{2.21} \\
  \bottomrule
  \end{tabular}
\vspace{-1em}
\end{table}

\begin{table}[tb]
  \caption{The results of various models on ImageNet $512 \times 512$. \textit{Pretrain} in this table denotes DiM trained at $256 \times 256$ resolution.}
  \label{tab:imagenet512}
  \centering
  \small
  \begin{tabular}{l|c|c|c|c}
 \toprule
 Mode & Images~(Iterations $\times$ BatchSize) & Parameters & GFlops & FID \\
 \midrule
 U-ViT-L/4~\cite{uvit} & 300M~(300K $\times$ 1024) & 287M &  76 & 4.67 \\
 U-ViT-H/4~\cite{uvit} & 500M~(500K $\times$ 1024) & 501M &  133 & 4.05 \\ 
 DiT~\cite{dit}  & 768M  &  675M & 524 & 3.04 \\ 
 DiffuSSM-XL~\cite{diffussm} & 302M & 660M & 1066 & 3.41 \\
 \midrule
 \textbf{DiM-Huge} &  \textit{Pretrain} + 15M~(64K $\times$ 240) & 860M &  708 &  3.94 \\
 \textbf{DiM-Huge} &  
 \textit{Pretrain} + 26M~(110K $\times$ 240)  & \text{860M}  &  \text{708}  &  \text{3.78}  \\
 \bottomrule
 \end{tabular}
\vspace{-1em}
\end{table}

\subsection{Quantitative Results}

\noindent
\textbf{ImageNet $256 \times 256$ pretraining.}
We compare DiM to other transformer-based and SSM-based diffusion models in~\cref{tab:imagenet256}. 
After training on 319 million image samples, DiM-Huge can achieve a score of 2.40 on FID-50K.
In the case that we use \textbf{63\%} of the training data of U-ViT~\cite{uvit}~(319M versus 500M), the performance of our model is comparable to the other transformer-based diffusion models, \ie, only about 0.1 worse on FID-50K.
When we train the model with 480M image samples, our model can outperform other models, achieving a score of \textbf{2.21} on FID-50K.
Moreover, compared to the DiffuSSM-XL, the Gflops of our Mamba-based diffusion model is much smaller, \ie, DiM requires fewer resources for inference. 

\noindent
\textbf{ImageNet $512 \times 512$ finetuning.}
Training on $512 \times 512$ image samples requires significant computational resources. Also, such a large resolution creates non-negligible latency during training and inference, shown in~\cref{fig:infer_speed}.
Therefore, instead of training from scratch, we finetune DiM-Huge pretrained on ImageNet $256 \times 256$ for higher resolutions, and report the results in~\cref{tab:imagenet512}.
With \textbf{3\%} of the $512 \times 512$ training data of U-ViT~\cite{uvit}~(15M versus 500M), DiM-Huge achieves 3.94 FID-50K. If we further finetune our model with 110K iterations, it can achieve 3.78 FID-50K.
The finetuned DiM-Huge can produce visually appealing $512 \times 512$ images, shown in~\cref{fig:imagenet512cfg}.

\noindent
\textbf{CIFAR-10.} We compare our method to other diffusion models on CIFAR-10 dataset, and we report the results in~\cref{tab:cifar}. The results show that our method can have comparable performance to other methods with the similar number of parameters.

\subsection{Ablation Studies}

\begin{wraptable}{r}{21em} 
\setlength{\tabcolsep}{13pt}
  \caption{Ablation studies on architecture at CIFAR-10. \texttt{MS} denotes multiple scan patterns. \texttt{LSC} denotes the long skip connection. \texttt{PT} denotes the padding tokens. \texttt{LFE} denotes the local feature enhancement.}
  \vspace{-.5em}
  \label{tab:ablation_cifar}
  \centering
  \small
  \begin{tabular}{cccc|c}
 \toprule
  MS & LSC & PT & LFE & FID \\
 \midrule
  \checkmark & \checkmark & \checkmark & \checkmark & 2.92 \\ 
  \checkmark &  \checkmark & \checkmark &  & 3.07   \\ % 2, 3.87
  \checkmark  & \checkmark &   & \checkmark  & 2.94 \\ % 2 , 3.25
   \checkmark &  &  \checkmark & \checkmark & 6.23 \\
    & \checkmark &  \checkmark & \checkmark & 17.60 \\ 
  \bottomrule
  \end{tabular}
\end{wraptable}

We conduct ablation studies on the CIFAR-10 dataset~\cite{cifar}.
Following U-ViT~\cite{uvit}, our model is trained for unconditional generation, employing the VP scheduler as defined in~\cite{scorebase}. 
We report FIDs in~\cref{tab:ablation_cifar}, where the first row contains the result of the best performance model, and the performance in other rows corresponds to the models without certain components.
According to the results, when comparing the first and the last two rows, we find the multiple scanning directions contribute most to the performance, showing the importance of global receptive field.
We also find that the long skip connection are beneficial for the training convergence, which is consistent with the findings in~\cite{uvit}.
Moreover, our two convolutions and the padding tokens are also beneficial for the performance.

\begin{figure*}
    \centering
    \begin{subfigure}[h]{0.49\textwidth}
        \centering
        \includegraphics[width=1\linewidth]{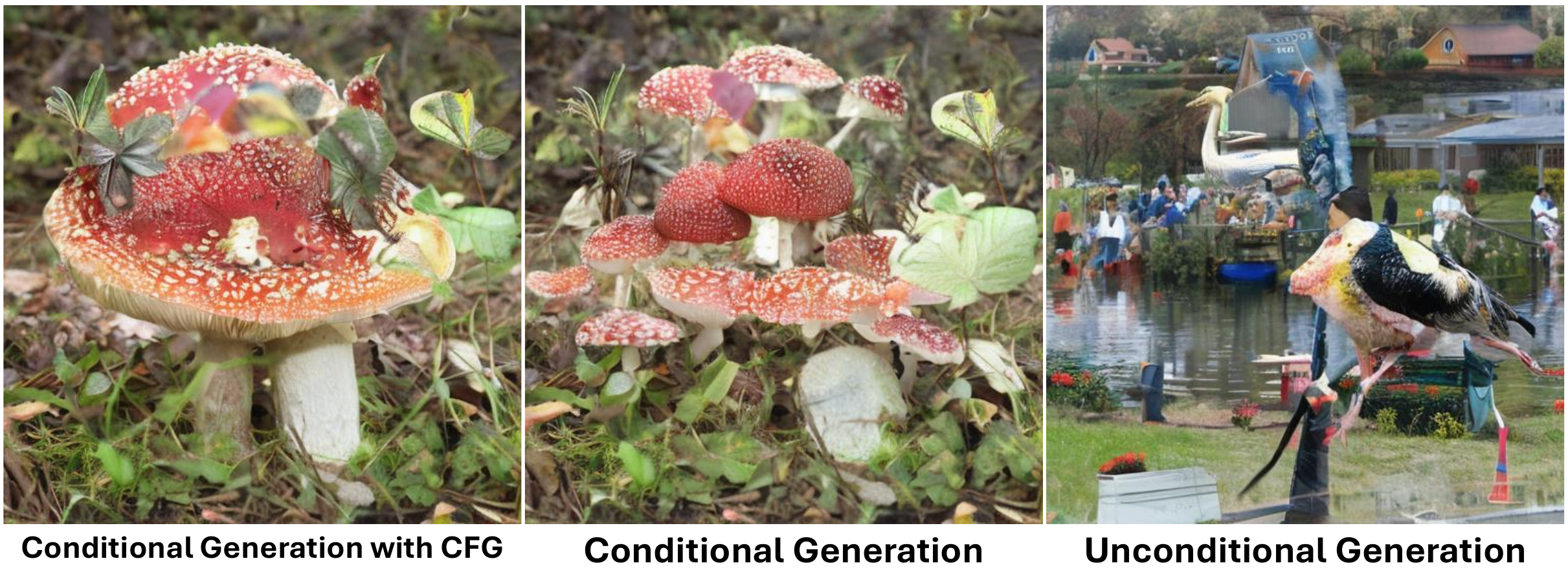} 
        \caption{
        Model with the scan patterns \ding{172}\ding{173}\ding{174}\ding{175}
        }
        \label{fig:abla_row_column_scan}
    \end{subfigure}
    \hfill
    \begin{subfigure}[h]{0.49\textwidth}
        \centering
        \includegraphics[width=1\linewidth]{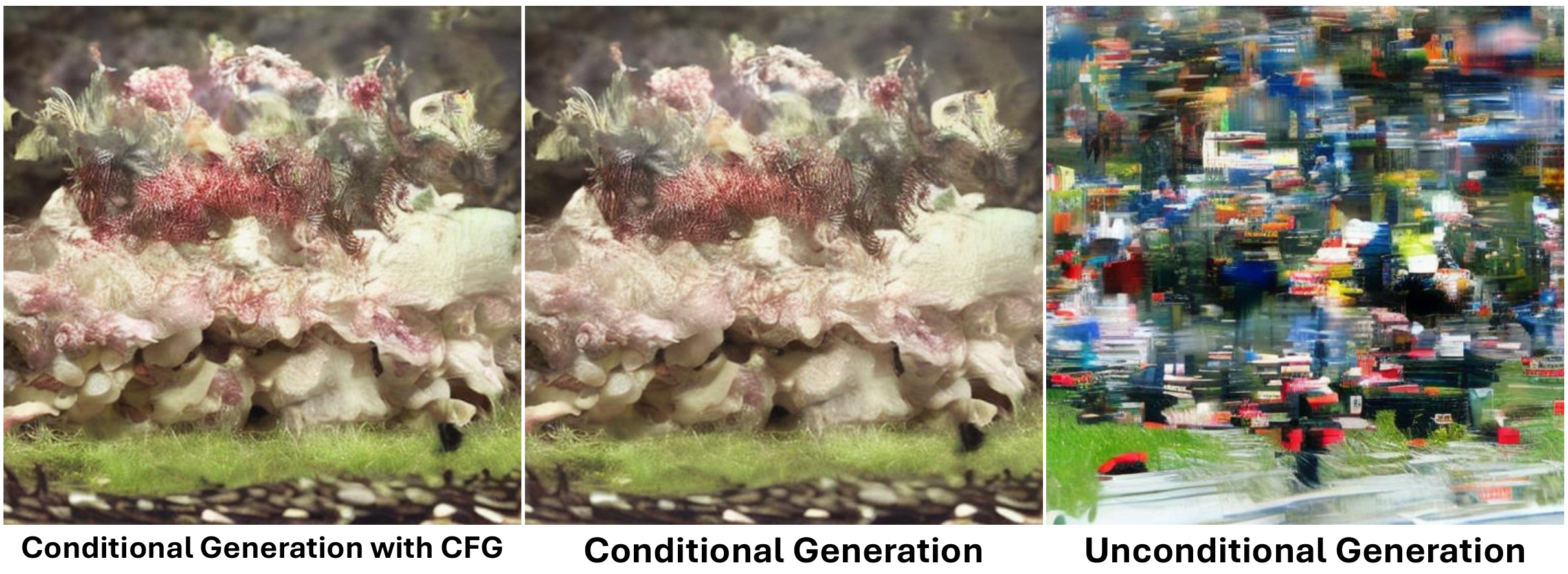}
        \caption{
        Model with the scan patterns \ding{172}\ding{173}
        }
        \label{fig:abla_row_scan}
    \end{subfigure}
\caption{
The \textbf{initial} $512 \times 512$ images generated by the model trained on $256 \times 256$ images without any additional techniques \textbf{before finetuning}. Comparing (a) and (b), we find DiM with row- and column-major scans can provide better object structures than the model with only row-major scans.
The circled numbers corresponds to~\cref{fig:main}.
}
\label{fig:direct512_init}
\vspace{-1em}
\end{figure*}

We further investigate different scanning patterns.
We find that using 2 or 4 directions doesn't influence the quantitative results of pre-training on low-resolution images, as shown in~\cref{tab:scans}.
However, according to the results in~\cref{fig:direct512_init}, the model combining row-major and column-major scan can produce a better initialization for higher-resolution generation, so we use four scans (presented in~\cref{fig:main}) in our model.

\begin{table}[t]
\parbox{.47\linewidth}{
  \setlength{\tabcolsep}{13pt}
  \caption{Benchmark of unconditional image generation on CIFAR-10~\cite{cifar}.}
  \vspace{.2em}
  \label{tab:cifar}
  \centering
  \small
  \begin{tabular}{l|c|c}
 \toprule
 Mode  & Parameters  & FID \\
 \midrule
 DDPM~\cite{ddpm} & 36M & 3.17 \\
 GenViT~\cite{genvit} & 11M & 20.20 \\
 U-ViT-S~\cite{uvit} & 44M & 3.11 \\ 
 \textbf{Ours-S} & 50M & 2.92 \\ 
 \bottomrule
 \end{tabular}
}
\hfill
\parbox{.47\linewidth}{
\setlength{\tabcolsep}{13pt}
  \caption{Ablation Studies of the scanning patterns on CIFAR-10~\cite{cifar}. The circled numbers corresponds to~\cref{fig:main}.}
  \label{tab:scans}
  \centering
  \small
  \begin{tabular}{c|c}
    \toprule
  Scanning Patterns  & FID \\
    \midrule
 \ding{172}\ding{173}\ding{174}\ding{175}  & 2.92 \\
 \ding{172}\ding{173}  & 2.91 \\
 \ding{172} & 17.60 \\ 
  \bottomrule
  \end{tabular}
}
\vspace{-1em}
\end{table}

\section{Conclusion}

In this paper, we propose Diffusion Mamba~(DiM), a new Mamba-based diffusion model backbone, for efficient high-resolution image generation. 
In our framework, the sequence model Mamba is used to process the patch features of the two-dimensional noisy inputs.
To adapt Mamba for two-dimensional data, we propose several approaches, including scan pattern switching, learnable padding token, and lightweight local feature enhancement.
Then, to efficiently train the model with high-resolution image samples, we propose to use a ``Weak-to-strong'' training and fine-tuning for DiM.
The experiments demonstrate that our model can achieve comparable performance with other transformer-based diffusion models on high-resolution image generation. We also explore the traning-free upsampling of DiM to generate higher-resolution images without further finetuning.
We have also add the failure cases, limitations and broader impacts in our appendix.

{
\small
\bibliographystyle{unsrt}
\bibliography{main}
}

\newpage
\appendix

\noindent

\section{More Results on ImageNet}

\begin{figure}[h]
    \centering
    \includegraphics[width=0.9\linewidth]{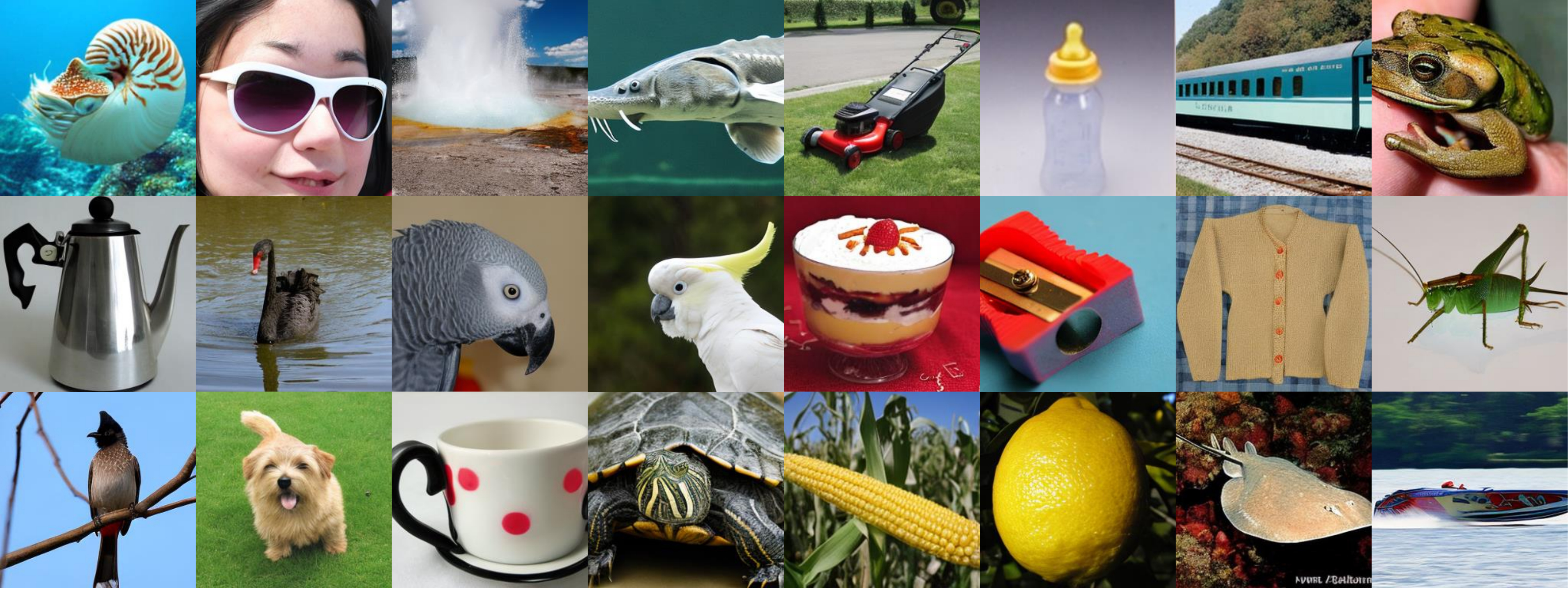}
    \caption{More $256 \times 256$ images generated by DiM-Huge pretrained on ImageNet with \texttt{cfg=4.0}.}
    \label{fig:imagenet256cfg4more}
\end{figure}

\begin{figure}
    \centering
    \includegraphics[width=0.99\linewidth]{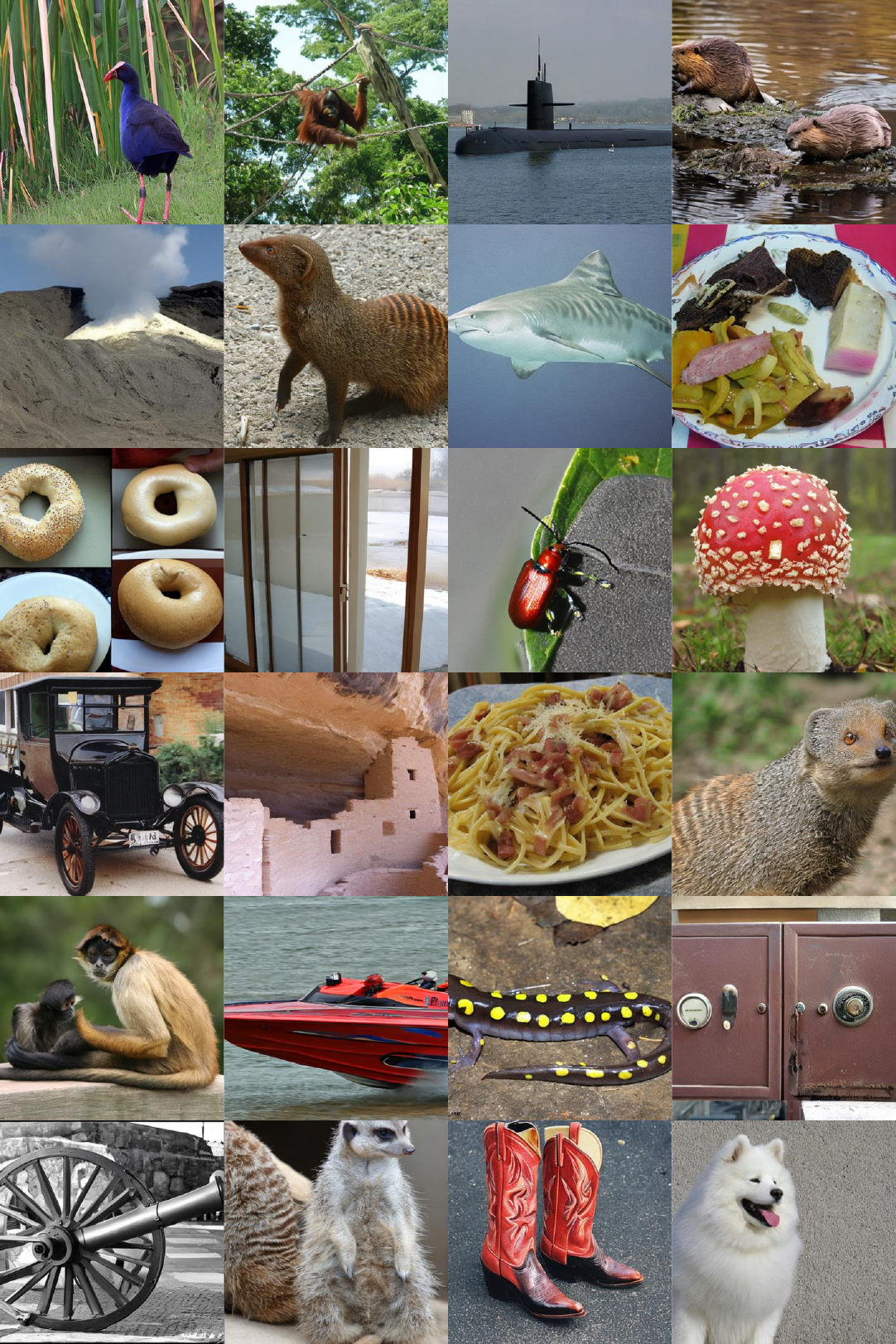}
    \caption{More $512 \times 512$ images generated by DiM-Huge finetuned on ImageNet with \texttt{cfg=4.0}.}
    \label{fig:imagenet512cfg4more}
\end{figure}

\begin{figure}
    \centering
    \includegraphics[width=0.99\linewidth]{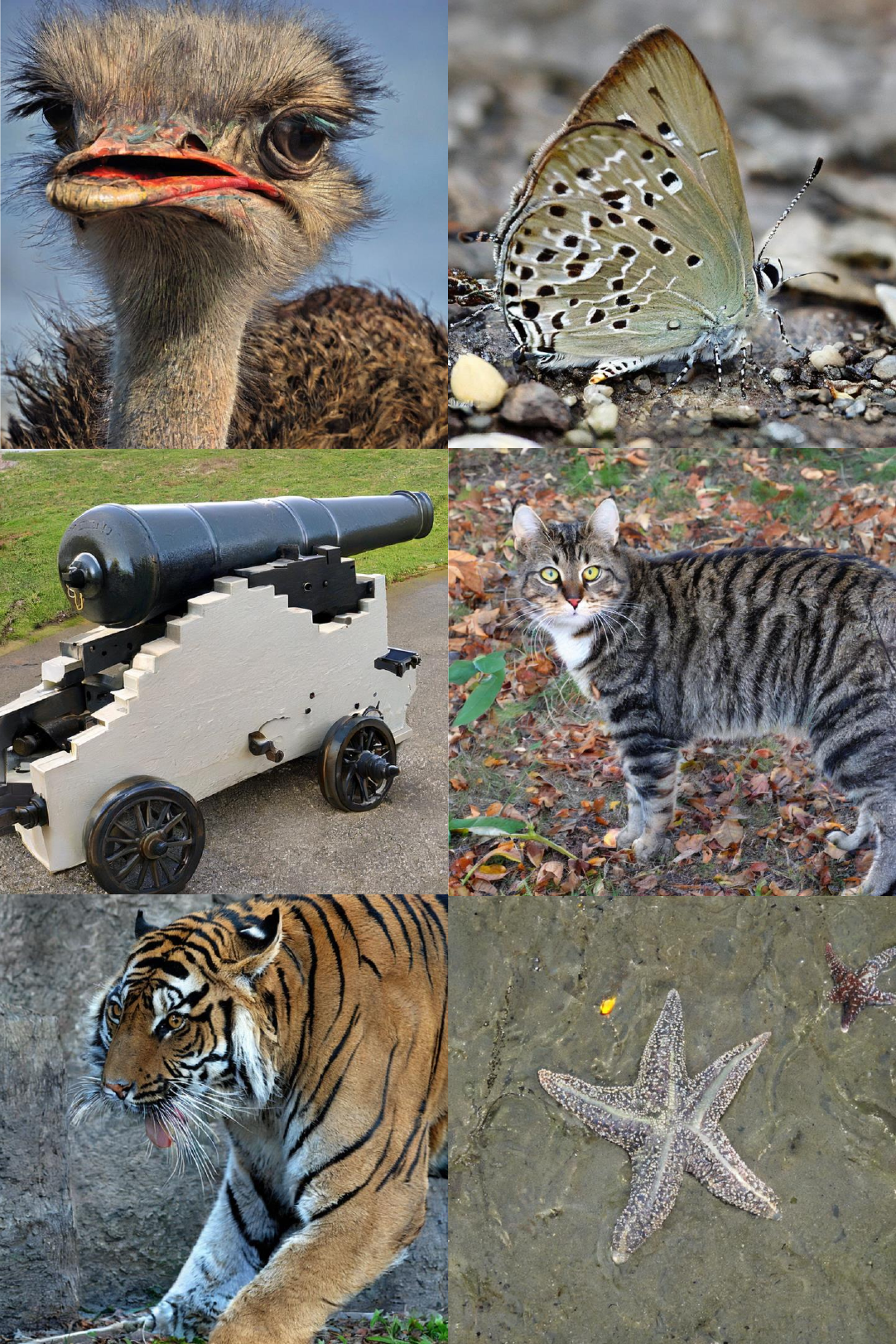}
    \caption{More $1024 \times 1024$ images generated by DiM-Huge trained with $512 \times 512$ images.}
    \label{fig:imagenet1024more}
\end{figure}

\begin{figure}
    \centering
    \includegraphics[width=1.\linewidth]{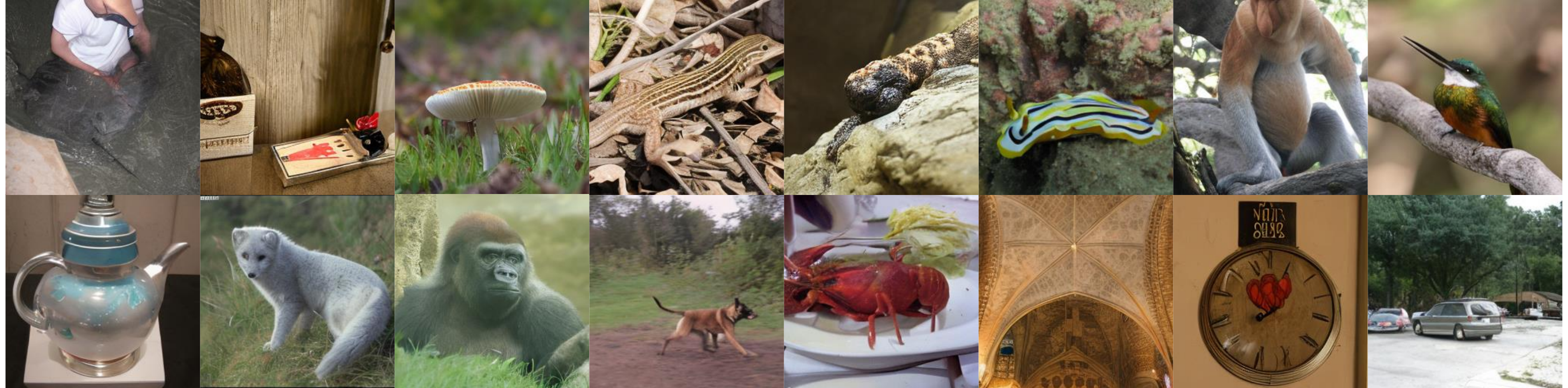}
    \caption{$256 \times 256$ images generated by DiM-Huge pre-trained on ImageNet with \texttt{cfg=1.4}.}
    \label{fig:imagenet256cfg_14}
\end{figure}

\begin{figure}
    \centering
    \includegraphics[width=1.\linewidth]{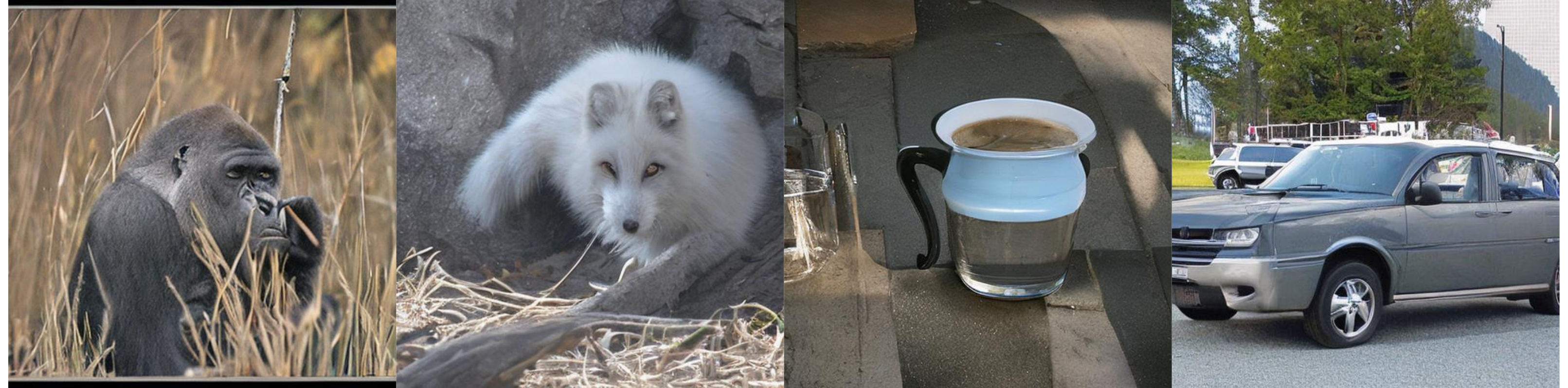}
    \caption{$512 \times 512$ images generated by DiM-Huge finetuned on ImageNet with \texttt{cfg=1.7}.}
    \label{fig:imagenet512cfg_17}
\end{figure}

We present more results of DiM trained on ImageNet with high classifier-free guidance, shown in~\cref{fig:imagenet256cfg4more},~\cref{fig:imagenet512cfg4more} and~\cref{fig:imagenet1024more}.
Notably, the classifier-free guidance has a great impact the visual quality and FID. The larger guidance weight improves the visual quality but leads to poor FID.
Empirically, the best classifier-free guidance weight for FID is set to \texttt{1.4} and \texttt{1.7} for ImageNet $256 \times 256$ and $512 \times 512$, respectively. We also present the generated images under these settings in~\cref{fig:imagenet256cfg_14} and~\cref{fig:imagenet512cfg_17}.

\begin{figure}
    \includegraphics[width=1\linewidth]{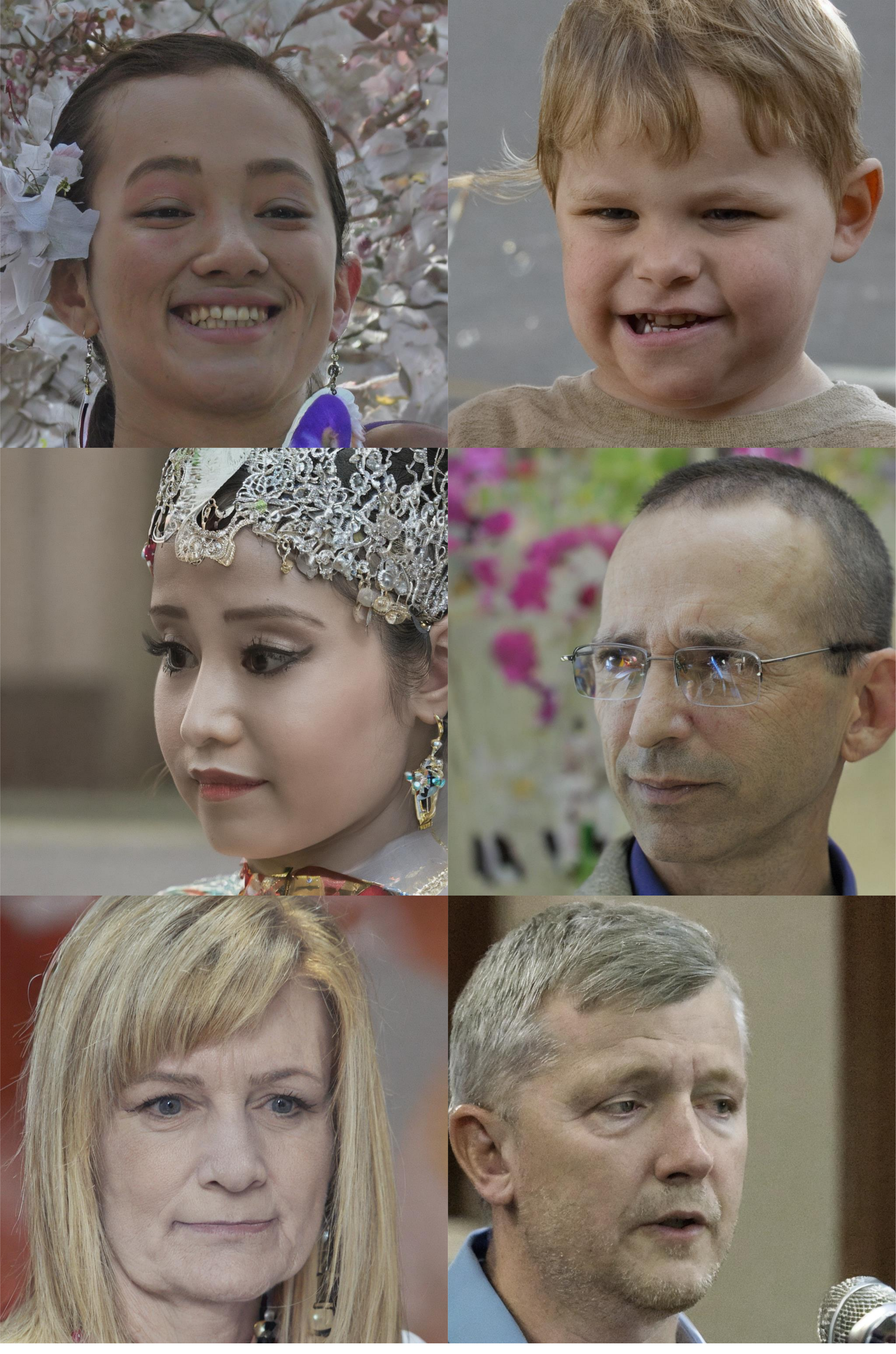}
    \caption{FFHQ $1024 \times 1024$}
    \label{fig:ffhq1024}
\end{figure}

\begin{figure}
    \begin{subfigure}[h]{0.39\textwidth}
        \includegraphics[width=1\linewidth]{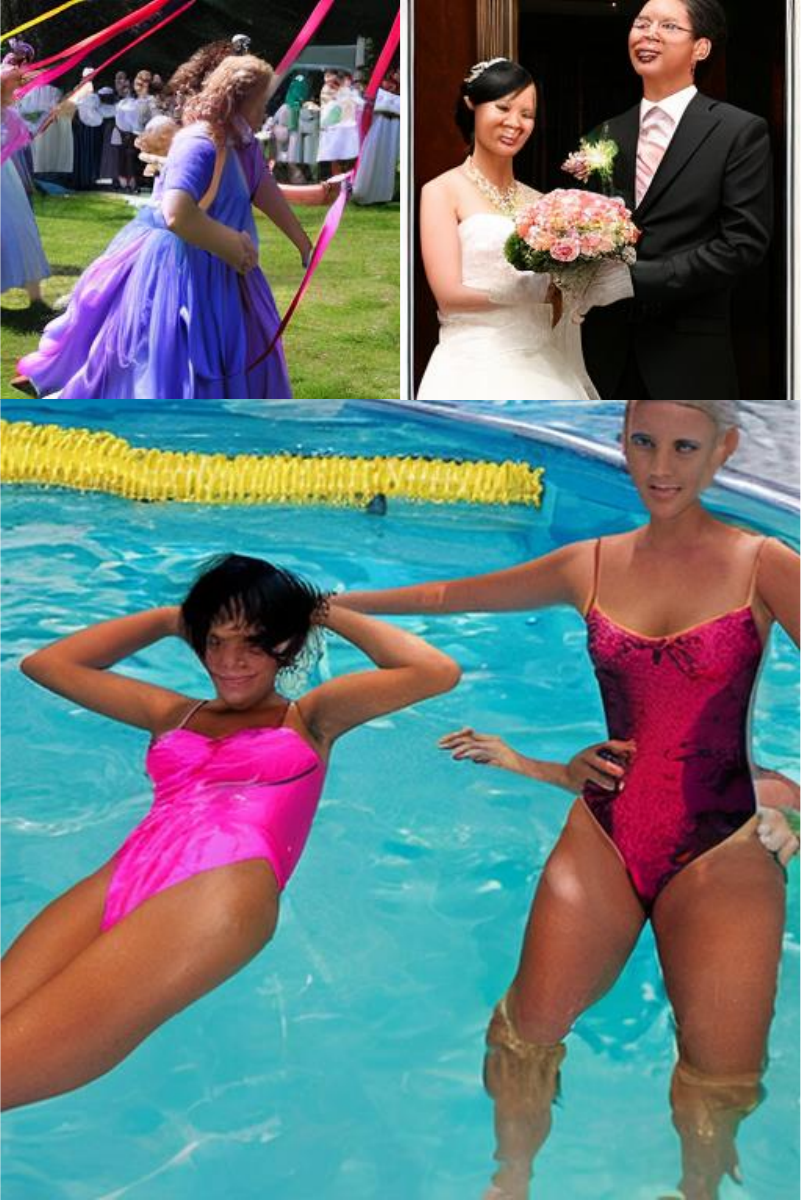}
        \caption{Failure cases on human.}
        \label{fig:fail_human}
    \end{subfigure}
    \begin{subfigure}[h]{0.59\textwidth}
        \includegraphics[width=1\linewidth]{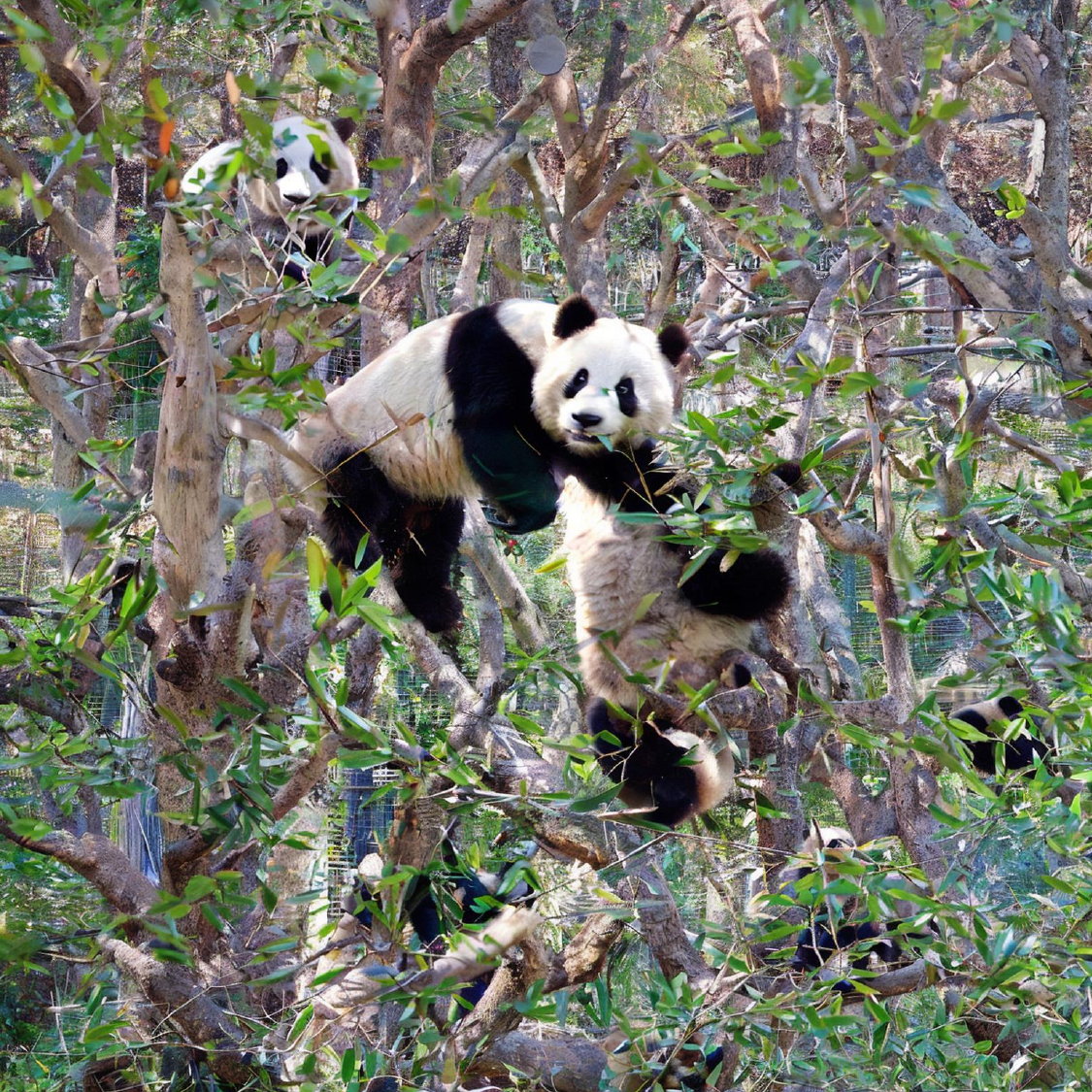}
        \caption{Failure cases on the training-free up-sampling.}
        \label{fig:failure_multi_highfreq}
    \end{subfigure}
\caption{
Failure cases. 
(a)~At each resolution, the image quality on human cases is unstable. The generated human faces and limbs are easy to collapse.
(b)~the problem of repeating patterns is not well resolved by the training-free upsampling at the resolution of $1536 \times 1536$.
Also, the background becomes cluttered, illustrating that DiM trained at lower resolution still has difficulty processing the details at $3 \times$ higher resolution.
}
\label{fig:failure}
\end{figure}

\section{Experiments on FFHQ}

\noindent
\textbf{FFHQ Dataset.}
We train DiM-Large for unconditional image generation. To verify the image generation ability of our model on more than \textbf{16K} tokens~(\ie, Mamba is faster than Transformer when the number of tokens greater than 10K), we set patch size as $\mathbf{1 \times 1}$ for FFHQ $1024 \times 1024$ dataset~(the number of tokens is calculated by $(\frac{1024}{1 \times 8})^2 = 16,384$).

\noindent
\textbf{Training Details.}
We first train DiM on FFHQ $256 \times 256$ with a batch size of 1024 for 50K iterations. Then, we finetune this pretrained model on FFHQ $512 \times 512$ for 20K iterations with a batch size of 256. Last, we further finetune the model on FFHQ $1024 \times 1024$ for 50K iterations with a batch size of 64. Note that the batch size on each resolution is different, so we use the similar number of iterations for finetuning.

\noindent
\textbf{Quantitative Results.} We also show the results of DiM trained on FFHQ $1024 \times 1024$ with $1 \times 1$ patch size. The generated images shown in~\cref{fig:ffhq1024} reveal that our DiM has the capability of processing on \textbf{16K} patches, and generate high-resolution human faces after training.

\section{Limitation and Failure Cases}
\label{sec:limitation}

According to the generated images in~\cref{fig:failure}, DiM still has limitations on generating high-resolution images.
First, at each resolution, the image quality on human cases is unstable. The generated human faces and limbs are easy to collapse. Specifically, in~\cref{fig:fail_human}, the facial features are not well positioned. This illustrates that DiM still has trouble in learning the details of images.
Second, the problem of repeating patterns is not well resolved by the training-free upsampling at $3 \times$ higher resolution. In~\cref{fig:failure_multi_highfreq}, the body parts of giant pandas appear in many places.
Also, the background~(the bamboo forest) becomes to messy, illustrating that DiM trained at lower resolution still has difficulty processing the details at $3 \times$ higher resolution.

\section{Broader Impacts}
\label{sec:broader_impacts}
Image generation has wide applications in assisting users, designers, and artists in creating new content. However, researchers, developers, and users should also be aware of the potential negative social impact of image generation models. They might be misused for generating misleading content and biased content.

\end{document}